\documentclass{article}
\usepackage{arxiv}
\usepackage{float}
\usepackage{verbatim} 
\usepackage{subcaption}
\usepackage{multirow}
\usepackage{amsmath,amssymb,amsfonts}
\usepackage[title]{appendix}
\usepackage[ruled, vlined,linesnumbered]{algorithm2e}
\restylefloat{figure}
\restylefloat{table}
\usepackage[utf8]{inputenc} 
\usepackage[T1]{fontenc}    
\usepackage{hyperref}       
\usepackage{url}            
\usepackage{booktabs}       
\usepackage{amsfonts}       
\usepackage{nicefrac}       
\usepackage{microtype}      
\usepackage{lipsum}
\usepackage{graphicx}
\usepackage{xcolor}
\usepackage{csquotes}
\usepackage{longtable}
\usepackage{pdflscape}
\usepackage{multirow}
\usepackage{tcolorbox}
\graphicspath{{media/}}     
\title{Assisting Clinical Practice with\\Fuzzy Probabilistic Decision Trees}

\author{
  Emma L. Ambags \\
  Department of Industrial Engineering \& Innovation Sciences, Eindhoven University of Technology, \\ 
  Eindhoven, The Netherlands \\
  \texttt{emma@ambags.com} \\
   \And
Giulia Capitoli \\
  School of Medicine and Surgery, University of Milano-Bicocca, Monza, Italy  \\ 
  Bicocca Bioinformatics, Biostatistics and Bioimaging (B4) research center, Milan, Italy \\
  \texttt{giulia.capitoli@unimib.it} \\
\And
Vincenzo L'Imperio \\
  Department of Medicine and Surgery, Pathology, University of Milan-Bicocca, \\ IRCCS Fondazione San Gerardo dei Tintori, Monza, Italy \\
  \texttt{vincenzo.limperio@unimib.it} \\
\And
Michele Provenzano \\
  Nephrology, Dialysis and Renal Transplant Unit, IRCCS—Azienda Ospedaliero-Universitaria di Bologna, \\ Alma Mater Studiorum University of Bologna, 40126 Bologna, Italy \\
  \texttt{michiprov@hotmail.it} \\
  \And
Marco S. Nobile \\
  Department of Environmental Sciences, Informatics and Statistics (DAIS), \\ Ca' Foscari University of Venice, Venice, Italy \\
  \texttt{marco.nobile@unive.it} \\
  \And
Pietro Li\`o \\
  Department of Computer Science and Technology, University of Cambridge, Cambridge, UK \\
  \texttt{pl219@cam.ac.uk} \\
}

\begin{document}
\maketitle

\begin{abstract}
The need for fully human-understandable models is increasingly being recognised as a central theme in AI research. The acceptance of AI models to assist in decision making in sensitive domains will grow when these models are interpretable, and this trend towards interpretable models will be amplified by upcoming regulations. 
One of the killer applications of interpretable AI is medical practice, which can benefit from accurate decision support methodologies that inherently generate trust.
In this work, we propose FPT, (MedFP), a novel method that combines probabilistic trees and fuzzy logic to assist clinical practice. This approach is fully interpretable as it allows clinicians to generate, control and verify the entire diagnosis procedure; one of the methodology's strength is the capability to decrease the frequency of misdiagnoses by providing an estimate of uncertainties and counterfactuals. Our approach is applied as a proof-of-concept to two real medical scenarios: classifying malignant thyroid nodules and predicting the risk of progression in chronic kidney disease patients. Our results show that probabilistic fuzzy decision trees can provide interpretable support to clinicians, furthermore, introducing fuzzy variables into the probabilistic model brings significant nuances that are lost when using the crisp thresholds set by traditional probabilistic decision trees. We show that FPT and its predictions can assist clinical practice in an intuitive manner, with the use of a user-friendly interface specifically designed for this purpose. Moreover, we discuss the interpretability of the FPT model.
\end{abstract}


\section{Introduction}
\label{introduction}
A Clinical Decision Support System (CDSS) is a health information system that helps healthcare providers make decisions in order to improve patient care. These systems provide assistance in the complex decision making processes of clinicians, by offering targeted clinical knowledge, care plan recommendations and other relevant health information at the point of care. With the extraordinary growth of biomedical data the need for a computational basis of medical practice is naturally emerging. Having AI systems alongside human experts has been found to increase the accuracy and efficiency in seizure detection \cite{yang2022continental}.
Occasionally these systems outperform human experts, as they mimic reasoning by medical professionals, but faster and less prone to human errors. As a result patient care can be improved whilst simultaneously practice variability is reduced. 

In spite of the proven success of Machine Learning (ML) and Artificial Intelligence (AI) algorithms in several other disciplines, the acceptance of such models to assist in highly sensitive domains (e.g., clinical environments) is still limited due to trust and transparency issues \cite{ahmad2018interpretable}. In the clinical field, black boxes are unacceptable \cite{shortliffe2018clinical}, for the clinical field is a sensitive domain in which the healthcare provider needs to be able to justify the rationale behind any decision. The Black Box Problem is a known shortfall of AI and ML based decision support systems; it refers to a system or program allowing the user to see its input and output, but is useless in explaining how it came to its output. Therefore, it is important that the basis of predictions and recommendations that are offered by a CDSS are understandable and interpretable to its users, improving the trust and acceptance of the generated suggestions. 

The need for transparency and interpretability is increasingly being recognised as a central theme to be addressed by AI research, especially when it operates in healthcare \cite{cabitza2017unintended, nguyen2022attentive}. 
Even more so due to upcoming regulations. Such as the General Data Protection Regulation (GDPR), as imposed by the European Union (EU) in May 2016. 
Among other things these regulations are supposed to shield citizens from decision making based on black boxes \cite{EUGDPR}. 
Moreover, in April of 2021, the Commission of the EU published a draft of what is called the Artificial Intelligence Act (AIA) \cite{EUAIA}. 
The proposed rules are specifically aimed at regulating the development and use of AI. 
These proposed regulations impose AI systems to comply with a set of mandatory requirements for trustworthy AI to be allowed to be placed on the EU market, primarily to enforce transparency and human oversight obligations to high-risk AI systems, e.g., safety-critical systems such as systems deployed in clinical environments (see also  \cite{rudin2019stop}). 
CDSSs specifically, as they operate in the sensitive clinical field, will be subject to the strictest set of requirements \cite{EUAIA}. Therefore, in order to effectively support the decision making process, computational methods need to be explainable and interpretable \cite{kundu2021ai}. 
This will allow for the decision making process of humans to be supported by a computational model. 
This interplay of human and AI meets the guidelines of automated decision-making as prescribed in the General Data Protection Regulation \cite{EUGDPR} and it will be able to meet the regulations as proposed in the Artificial Intelligence Act (AIA) by the European Union (EU) \cite{EUAIA}. 

The potential of CDSSs is evident and the need for these systems to be interpretable is ever growing, leading to hundreds of works exploring different approaches   \cite{hakkoum2021interpretability}. 
In practice, end-users (i.e., healthcare providers) of CDSSs are less likely to trust the recommendations of systems whose workings they do not understand \cite{burrell2016machine}. 
Moreover, CDSSs will be challenged legally by having to comply with future requirements as imposed by the AIA. 
Therefore, this research focuses on interpretable AI (IAI) rather than explainable AI (XAI). 

XAI methods use \textit{post hoc} analysis of the decision making process, so that it provides insights in the way decisions have been taken by the model. 
Thus, XAI only peers inside the model after it has been created. Concretely, an XAI model will first involve building a black box model, after which it will help to dissect the internal mechanics of the black box model to understand the importance of various features and the decisions it can lead to \cite{bacsaugaouglu2022review}.
IAI creates models that are \textit{a priori} interpretable by humans, i.e., human-interpretable from the beginning \cite{fuchs2020pyfume}. 
The decision-making process of the system is directly observable. 
Interpretability is considered a broader term than explainability \cite{bacsaugaouglu2022review}. An interpretable system is where the user can not only see but also understand how inputs are mathematically mapped to outputs \cite{doran2017does}.

This paper presents an interpretable AI method that integrates fuzzy logic and probabilistic trees. Probability trees (PT) have highly self-explanatory semantics, being able to accurately represent and structure a decision problem \cite{raiffa1968decision}. 
The presentation of a PT is descriptive and simple to understand, and intuitively visualises conditional probabilities that build on Bayes' theorem \cite{pearl2000models}.
Furthermore, the importance of evaluating uncertainty and vagueness in medical variables is stressed. 
Think of ill-defined concepts such as: 'young' \textit{vs} 'old', 'small' \textit{vs} 'large', or a concept such as having a 'high fever' at a body temperature of 39 \textdegree C (but what about a temperature of 38.9\textdegree?). The theory of fuzzy sets, as introduced by \cite{zadeh1965fuzzy}, gives the possibility to formalise a partial membership of an element to a fuzzy set. Meaning that a variable can be in multiple states at once, to differing degrees. This concept allows for incorporating fuzzy relationships, and thus extending the possibilities of PTs. We build on the PT framework as presented in \cite{genewein2020algorithms}.

\begin{figure}[htb]
    \centering
    \includegraphics[width=\textwidth]{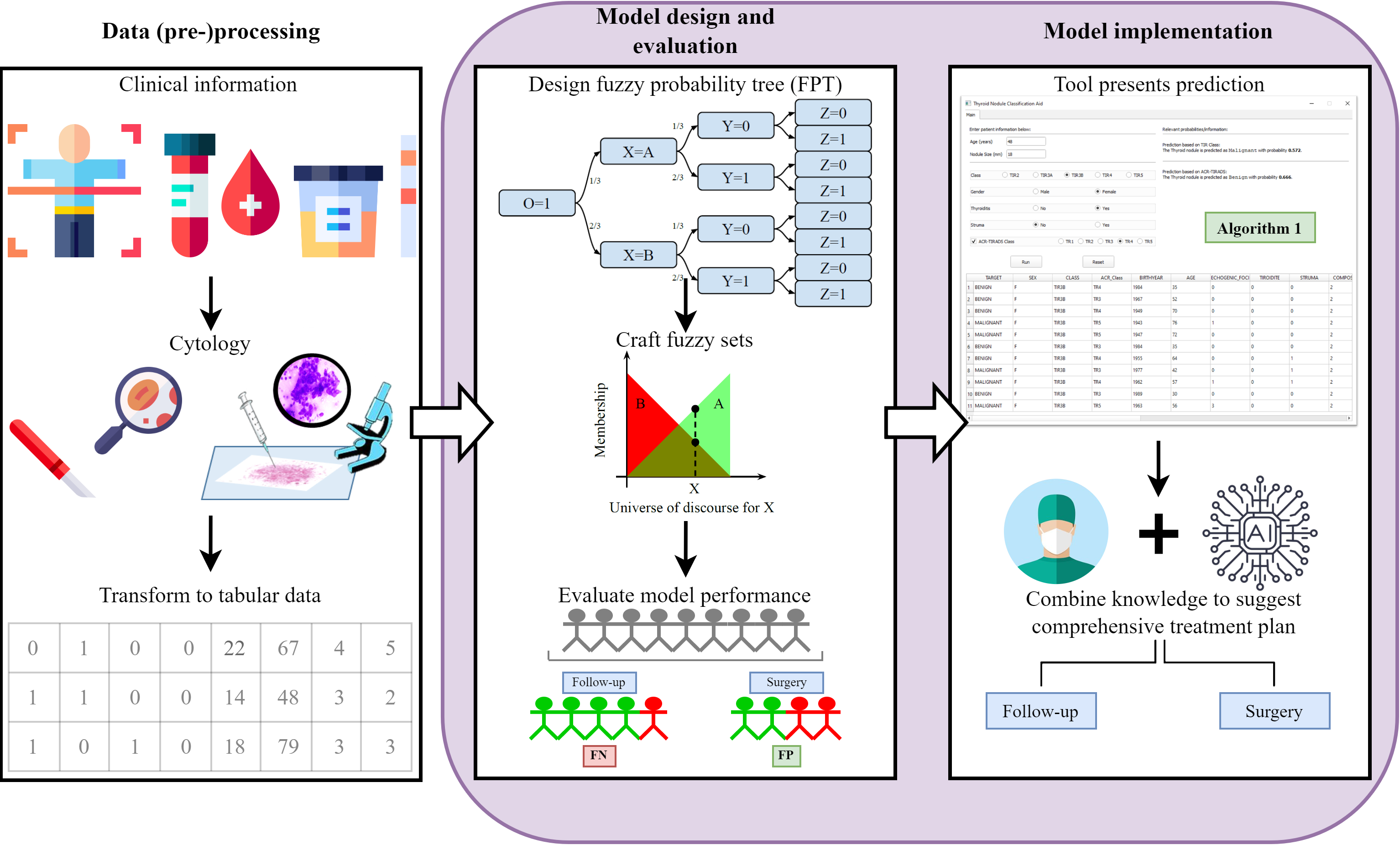}
    \caption[The summary of the approach taken in this project.]{Visual abstract of the approach taken within this research; modelling and decision support for clinical decision making. The first phase occurred outside this research, the data has been provided in tabular form. The phases in the purple rectangle have been carried out within this research. Firstly, the modelling phase is carried out (creation of the tree, crafting fuzzy sets, and assessing the performance). Secondly, a tool has been created to assist clinical decision making in an interpretable manner. The green label ``Algorithm 1'' indicates where in the decision making process the fuzzy probabilistic prediction algorithm acts.}
    \label{fig:summaryProject}
\end{figure}

In this work, we develop methods for a fuzzy logic driven probabilistic tree and apply it to assist in decision making in two clinical contexts, namely: (1) the classification of thyroid nodules, and, (2) determining the risk of patients progressing to the final and most harmful stage of chronic kidney disease over a 2-year period. Two different types of diseases are used for the case studies in order to test out and substantiate the generalisability of the FPT method across differing diseases. 
Figure \ref{fig:summaryProject} visualises the steps involved in each case study. The phases highlighted by the purple rectangle are executed within this research. The green label containing ``Algorithm 1'' in the column on the right indicates where the fuzzy probabilistic decision tree is used in the decision making process. 

The integration of probabilistic trees and fuzzy reasoning, leading to a hybrid tree, will provide an AI system that is aligned with the way humans reason. The performance of the fuzzy probability tree is compared to several other interpretable decision making models, namely regular PTs, Decision Trees, and Logistic Regression. In order to promote use-ability and interpretability a tool is created that provides an interactive and natural way to interact with the decision making algorithm. Considering the interpretable method, the human-like reasoning due to integrating fuzzy logic, and the naturally understandable tool clinical practice will be able to benefit from bringing AI-in-the-loop in daily decision making processes. In addition, the methods can be used to identify records of past patients with similar profiles based on the fuzzy probabilistic tree. An implementation that can provide very useful information when the industry increasingly leverages from data sharing.\\

This manuscript is organised as follows. Existing methods on which this paper builds are outlined in Section \ref{methods}. Section \ref{fpt} describes the proposed FPT decision making method. The two case studies used to assess the performance of FPT are described in Section \ref{casestudies}, this section also outlines the creation of the trees and the fuzzy sets, as well as providing a walk-through of an example showing how the algorithm makes predictions. Section \ref{results} presents the achieved experimental results of the FPT compared to other interpretable benchmark models. An analysis of the interpretability of FPT is presented in Section \ref{interpretability}. The user interface that has been designed as a tool for clinicians is discussed in Section \ref{tool}. Finally, discussions and conclusive remarks are presented in Sections \ref{discussion} and \ref{conclusion}, respectively. 

\section{Methods} \label{methods}
\subsection{Probabilistic Trees}

\begin{figure}[!hb]
    \centering 
    \includegraphics[width=.55\textwidth]{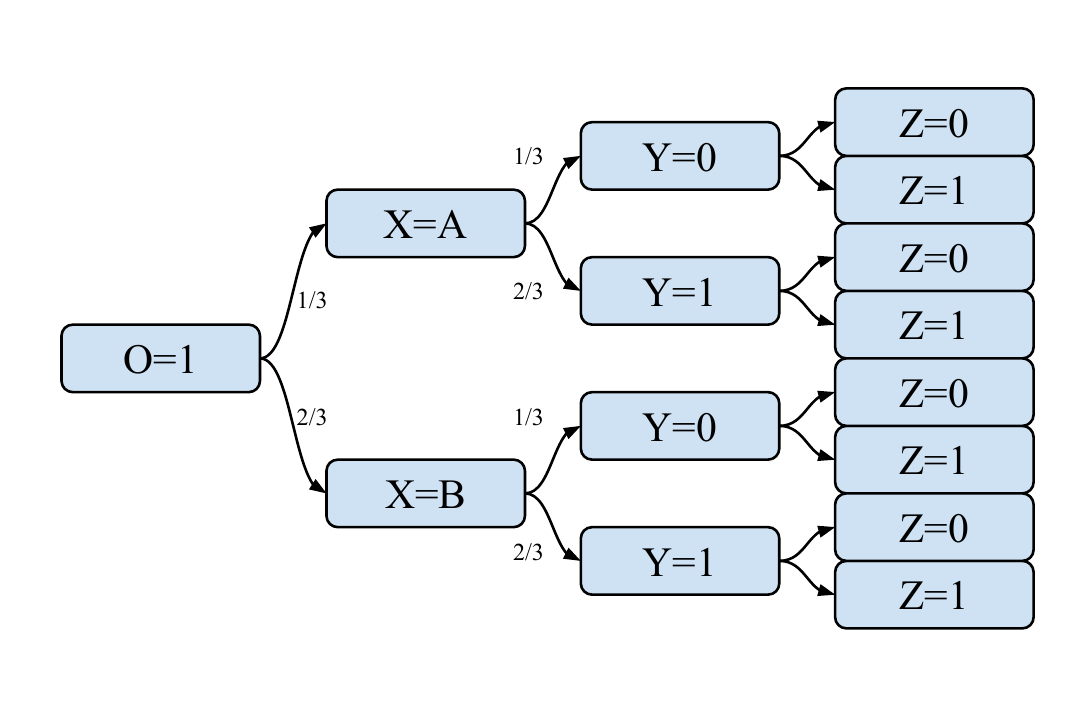}
    \caption{Scheme of a PT with 3 variables.}
    \label{fig:justpt}
\end{figure}

(Discrete) Probability Trees (PTs) are causal models that use nodes to represent the potential states of the processes, and the arcs between nodes indicate both the probabilistic transitions and the causal dependencies between them \cite{genewein2020algorithms}.
Similarly to Bayesian Networks (BNs), they can be used to model causal relationships and perform inference. 
However, thanks to the fact that they are not represented as a directed acyclic graphs, PTs allow to model multiple alternative scenarios where variables do not necessarily follow a partial order.
Formally, the nodes modelling the variables in a  PT are tuples $n=(u, \mathcal{S}, \mathcal{C})$ where: $u$ is a numeric identifier; $\mathcal{S}$ is a list of statements and $\mathcal{C}$ is a ordered set of transitions.
A statement has the form \texttt{X IS something},  which is technically represented as a tuple $(X, $ something$) \in \chi \times \mathcal{V}_x$ where $\chi$ is the set of variables and $\mathcal{V}_X$ is the range of the variable $X$.
Such range is sometimes well defined (e.g., concepts like `p53 mutation = True'), but this is often not the case in real world scenarios where variables are more fuzzy (e.g., concepts like `fever = High').
Of course, such fuzzy concepts can, in principle, be converted to crisp concepts by using thresholds. 
This is a rather used approach  (e.g., `high fever' corresponds to a temperature $\geq 38$\textdegree) that unfortunately can introduce artefacts, it defines arbitrary partitions, and in general prevents us from properly handling uncertainty in our data (e.g., what about a temperature of 37.9\textdegree?). A \textit{(total) realisation} in the probability tree is a path from the root to a leaf, and its probability is obtained by multiplying the transition probabilities along the path. An \textit{event} is a collection of total realisations that we can filter using propositions about a random variable. Therefore, an event is used to describe a set of all realisations that contain a node with the specific statement, for instance the event '$X=A$' in Figure \ref{fig:justpt}. Furthermore, we can use logical connectives of negation (\texttt{Not}, $\neg$),
conjunction (\texttt{And}, $\land$) and disjunction (\texttt{Or}, $\lor$) \cite{genewein2020algorithms}. With this we can state composite events, such as '$\neg (X = 0 \land Y = 1)$'. 

One example of a simple PT with three variables is shown in Figure \ref{fig:justpt}.
\subsection{Fuzzy sets}
Human reasoning is able to process information expressed in vague and uncertain terms. 
The theory of fuzzy sets, introduced by \cite{zadeh1965fuzzy}, gives the possibility of formalising a partial membership of an element (of a universe of discourse) to a fuzzy set. 
Specifically, the membership degree $\mu(x)$ of an element $x$ can range from 0 (= no membership at all) to 1 (= full membership).

Fuzzy sets can be used to express linguistic terms that, in turn, can be combined to model  linguistic variables, in order to make the transition between fuzzy concepts smooth and as close as possible to human natural reasoning. As human reasoning often involves linguistic variables of ill-defined concepts.
This possibility can be exploited to express medical concepts that are fuzzy in nature  (e.g., a young \emph{vs} an elderly person).
We provide some examples of fuzzy sets applied to biomedical terminology in the Supplementary Material. 

A fuzzy set has gradual boundaries, in contrast to classical sets, which contain discrete borders.
The universe of discourse ($U$) is the set of possible values that a variable \textit{u} can take on. We can mathematically define a fuzzy set $\widetilde{A}$ as:

\begin{equation}
    \widetilde{A} = \{(u, \mu_{\widetilde{A}}(u)) \mid u \in U\}
\end{equation}
\\
Here, $\mu_{\widetilde{A}}(u)$ is the degree of membership of $u$ in the fuzzy set $\widetilde{A}$, assuming $\mu_{\widetilde{A}} (u) \in [0, 1]$.

\section{The proposed FPT} \label{fpt}
We propose the integration of PTs and Fuzzy set theory, therefore, calling the method Fuzzy Probability Tree (FPT). This allows for using the existing framework of PTs as developed in \cite{genewein2020algorithms}, while incorporating uncertainty about the data, or allowing for a flexible description of vague variables. Thus, enabling us to incorporate human expert knowledge in the form of fuzzy membership functions to probabilistic trees. 

While PTs require well defined discrete concepts and events, this is seldom the case in real world scenarios. Especially in the medical field, where the variables are often fuzzy, vague or ambiguous. By using the FPT approach -- and by means of carefully crafted fuzzy sets -- we can incorporate the inherent uncertainty in variables to balance the probabilities. The integration of PTs and fuzzy sets, leading to an FPT, will provide an AI method that is aligned with the way humans reason. Furthermore, (F)PTs are able to represent circumstances or \emph{explanations} that cannot be represented with other graphical techniques (e.g., Bayesian networks), paving the way to a novel form of interpretable AI.

Constructing the FPT is an iterative process, done in collaboration with domain experts. The selection of the features and the order of the features in the tree are based on domain knowledge (deduction), however, the transition probabilities in the trees are based on the data (induction). In making its predictions, the tree will present two probabilities derived from the FPT that has been constructed. It will present the probability associated with the data point belonging to the negative class (0: no disease) or positive class (1: disease), naturally these probabilities sum to 1.
By default we consider a cut-off at a probability of 0.50, which indicates that when the probability of the positive class is $\geq 0.50$ the model predicts the nodule as the positive class. By this we can transform the probabilities into binary predictions, which can be used to determine the overall performance of the probabilistic model. Although, part of the strength of the model lies in its ability to show the two probabilities, and with it provide its degree of certainty that the nodule is either predicted as negative or positive. This threshold can be modified to reflect the situation at hand. For example in times where prioritisation is required, the threshold can be increased to consider the more urgent/certain cases. 

\begin{algorithm}[ht]
\caption{predict (FPT)}
\label{predict}
\SetKwInOut{Input}{Input}
\SetKwInOut{Output}{Output}
\Input{tree: PT, node: tree node, statements: list, c\_statements: list, class: int }
\Output{probability for given class (for one patient)}
\If{node is leaf}{
return ratio of labels with requested class
}
\If{statements $= \varnothing$}{
return \textbf{conditionalProbability}(tree, \textbf{findExistingConditions}(tree, statements))}
next\_var, next\_val = statements$[0]$\\
probability = 0\\
\If{$\exists_{c \in \text{node.children}}$ c.var, c.val $==$ next\_var, next\_val}{
\textcolor{purple}{mf = \textbf{get\_membership}(c.var, c.val)}\\
\tcp{predict probability for this condition}
probability $+=$ \textcolor{purple}{mf} $\cdot$ \textcolor{teal}{\textbf{predict}}(tree, c, statements$[1:]$, c\_statements)}
\ElseIf{$\exists_{c \in \text{node.children}}$ c.var $==$ next\_var}{
\tcp{compute weighted probability over all children}
probability $+=$ \textcolor{purple}{mf} $\cdot$ \textbf{conditionalProbability}(tree, \textbf{findExistingConditions}(tree, c\_statements))}
\Else{
\textcolor{purple}{mf = $^1/_{|node.children|}$}\\
probability $=
\sum_{c \in node.children}{\text{\textcolor{purple}{mf}} \cdot \textcolor{teal}{\textbf{predict}}(\text{tree, c, statements, c\_statements})}$ }
return probability
\end{algorithm}

\begin{algorithm}[ht]
\label{findExistingC}
\caption{find existing conditions}
\SetKwInOut{Input}{Input}
\SetKwInOut{Output}{Output}
\Input{tree: PT, statements: list}
\Output{largest set of statements that exists in the PT: list}
current = tree.get\_root()\\
\For{idx in range(len(statements))}{
\tcp{Extract var, val pair for specific statement}
var, val = statements[idx]\\
\If{var != current.var}
{
return statements[:idx]
}
found = False\\
\For{ idx2 in range(len(current.children))}
{
label, child\_node = current.children[idx2]\\
\If {label == val}{
current = child\_node\\
found = True
}}
\If{not found}
{
return statements[:idx]
}

}
\end{algorithm}
Algorithm \ref{predict} takes as input the FPT, a pointer to a specific node of the FPT, a list of statements, and a specific class.  
The idea is to transverse the tree, characterise a specific patient with respect to the specified statements, and calculate the probability of the patient to belong to the given class. 

The algorithm begins by considering the base case of tree leaf (lines 1--2). 
In such a case, we simply count and return the number of cases for the requested class and terminate the execution. 

In lines 3--4, the algorithm checks whether the list of statements is empty. 
In this case, the algorithm uses the \texttt{conditionalProbability} helper function which calculates the probability of a specific class (e.g., malignancy) occurring given a set of conditions.
Else, the algorithm extracts the next statement, in the form of pair (variable,value), and checks whether there exists a child node corresponding to that specific statement. 
If so, the  fuzzy membership corresponding to this case is calculated (if the variable is not fuzzy, it is set to 1 by default). 
The membership is then multiplied by the probability of the rest of the sub-tree by means of  a recursive call (lines 7--9). 

If the pair in the statement was not found, the algorithm still tries to determine whether there is a child node corresponding to the variable. 
In this second case, it calculates the weighted probability across all children (lines 10-11).

Finally, if the variable does not exist as child, the algorithm calculates the probability of the chosen event by following all children, and weighs each prediction using a  membership value equal to 1 divided by the number of children nodes (lines 13--14). 


Furthermore, the algorithm uses the helper function \texttt{Find Existing Conditions}, this function will be called when a patient in the test set is not represented in the training set (i.e., there is no path in the tree to represent this patient). 
Then we wish to predict this patient using the weighted average of similar data points. 
In order to do so, this function returns a list of the largest set of conditions of this patient that is represented in the tree. 
In essence, the function finds the longest subset of statements that exist in the tree. 

\section{Two clinical case studies} \label{casestudies}
In this section the proposed FPT method is implemented to make predictions in two real medical scenarios. Firstly, a case of classifying malignant and benign thyroid nodules will be handled. Including detailed descriptions of the process of designing the tree, the inner workings of the tree, and setting out the differences between classifying using regular PTs and the FPT method. Thereafter, we discuss the case of predicting the 2-year risk of patients suffering from Chronic Kidney Disease (CKD) to progress to the most advanced stage of this disease, namely End Stage Renal Disease (ESRD), where patients typically undergo dialysis and experience kidney failure. The second case study will serve to substantiate the generalisability of the proposed methods. Furthermore, it is interesting to incorporate two unrelated case studies as they consider two different types of diseases, and the performance of the proposed methods can be examined more carefully. The two diseases are different in terms of the classifications introduced by the International Statistical Classification of Diseases and Related Health Problems (ICD). Thyroid cancer falls under the category of neoplasms (ICD-10 Code: C73) and CKD belongs to the diseases of the genitourinary system (ICD-10 Code: N18) \cite{who2019international}.
\subsection{A thyroid nodule case study}
Thyroid nodules are lumps within the thyroid gland, they are exceedingly common and a frequent find on neck sonography. As much as half of all people are found to have at least one thyroid nodule by the age of 60 \cite{frates2005management}. However,  only 5 to 10\% of thyroid nodules are found to be cancerous \cite{bessey2012incidence, ocal2019malignancy, smith2013risk}. How to distinguish between benign and malignant thyroid nodules is a great clinical challenge, one in need of solving in order to perform the appropriate surgery with the correct indication. Patients that undergo surgery to remove (part of) the thyroid gland, run the risk of needing lifelong hormone replacement therapy. However, many patients still needlessly undergo surgery in response to their thyroid nodules.
Which is burdensome for both the patient, as well as it leading to high healthcare costs for society. \\
Since the risks are high in the clinical field of predicting thyroid nodules, an effective yet interpretable model to assist in the complex decision making process of clinicians is considered very valuable. 

\subsubsection{The thyroid nodule data set}
This study includes real medical case data, including 448 patients who underwent Fine Needle Aspiration (FNA), guided by the United States (US) Thyroid Imaging Reporting and Data Systems (TIRADS). The patients were under treatment at the interventional radiology clinic, ASST Monza, Italy between the months of January and August of 2019 \cite{cancers13092230, PMID:34771602}. The original data set is of tabular form and contains the information on a total of 480 thyroid nodules that were subjected to FNA, each record represents a thyroid nodule case in a specific patient containing the relevant information for 33 features. Table \ref{tab:DataTypes} presents an overview of the relevant features and their types. From the 480 thyroid nodules, a total of 79 were excluded: 13 lesions with an unsatisfactory cytology and no FNA repetition; and 66 nodules with no histopathology/follow-up diagnosis available. Resulting in a total of 401 subjects to be considered in developing the model. Appropriate informed consent have been obtained from all patients. 

\begin{figure*}[htbp]
    \centering
    \includegraphics[width=\textwidth]{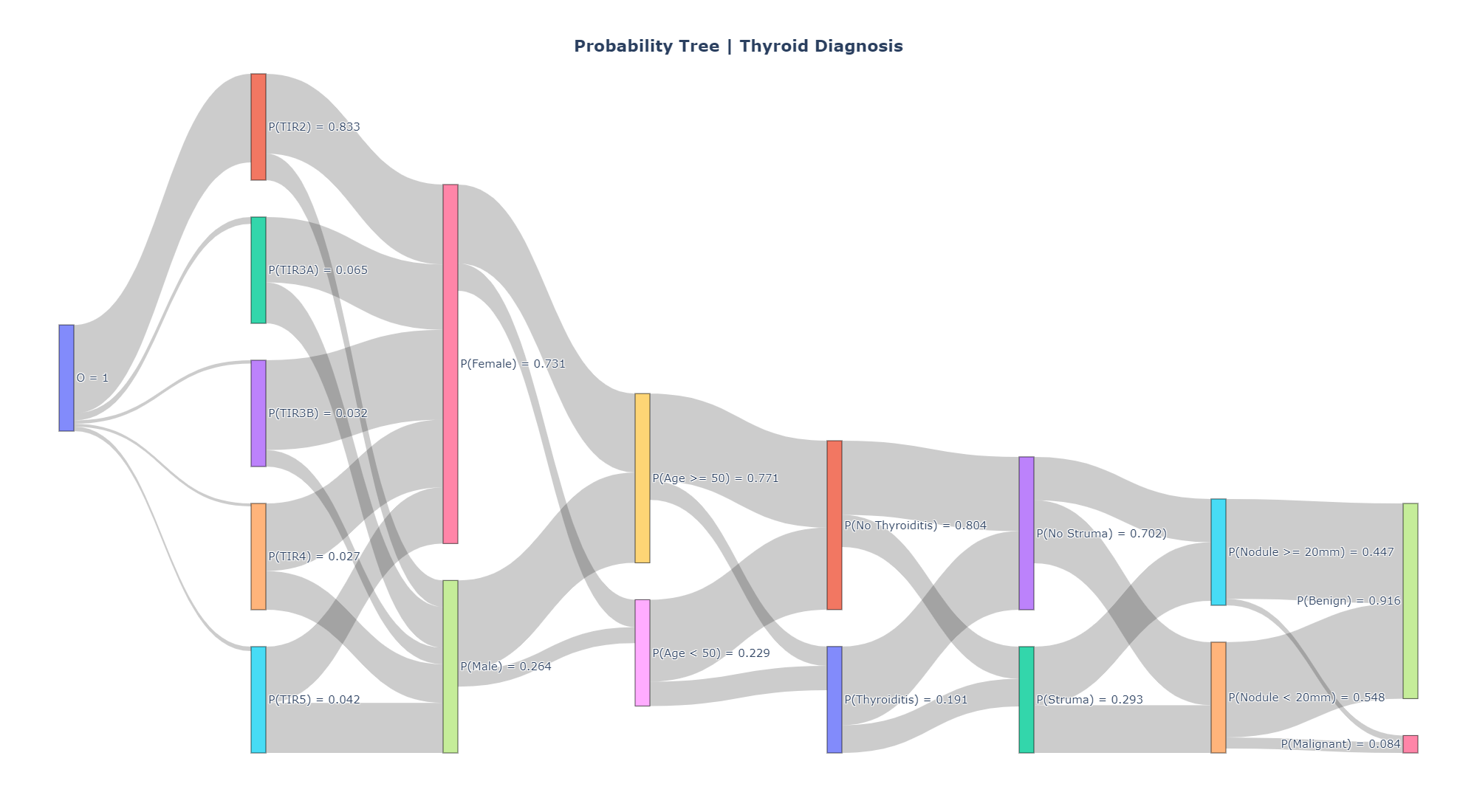}
    \caption{PT developed for the classification of thyroid nodules. Each arc represents the transition probability from one variable to the next, the size of the arcs are representative of the corresponding transition probabilities. The PT has been constructed based on the thyroid nodule data set. As a path is followed along the tree, it will end up in benign or malignant leaf nodes. The majority class in the leaf nodes will be the final prediction (in the case 'threshold $= 0.5$' is used).}
    \label{fig:probabtree}
\end{figure*}

A schematisation of the probabilistic tree created for this case study is visualised in Figure \ref{fig:probabtree}. It shows the possible values for all variables in the model, the arcs represent the conditional probabilities for the entire data set. This figure does not show every possible total realisation as a separate path in the probability tree, as we can see in the example handled in Figure \ref{fig:ptvsptfis}. Visualising the complete tree will blow up the image, due to the large number of paths. The tree as presented in Figure \ref{fig:probabtree} will -- on average -- have 2.5 data points in each total realisation up to a leaf (a total realisation is a path from root the leaf). Although in reality, some patient profiles are more common and the data points tend to clump together while others may not even occur at all. The average number of data points per total realisation is determined by calculating the possible number of combinations of feature values and dividing the total number of data points by this number ($\frac{401}{5 \cdot 2^{5}} = 2.51$).
\begin{table}[htbp]
\centering\small
\caption{Explanation of relevant features.}
 \begin{tabular}{||c | c||} 
 \hline
 \textbf{Description} & \textbf{Type} \\ [0.5ex] 
 \hline\hline
Date of birth & Ordinal \\ 
 \hline 
Gender & Binary \\
 \hline 
Date of FNA & Ordinal\\
 \hline
Thyroid suspiciousness class & Ordinal \\
 \hline
Final classification & Ordered \\
 \hline
Thyroiditis indication & Binary \\
 \hline
Struma indication & Binary \\
 \hline
Composition of thyroid nodule & Categorical \\
 \hline
Echogenicity of thyroid nodule & Categorical \\
 \hline
Shape of thyroid nodule & Categorical \\
 \hline
Margins of thyroid nodule & Categorical \\
 \hline
Measure presence of echogenic foci & Categorical \\
 \hline
EU TIRADS indication & Categorical \\
 \hline
ACR TIRADS indication & Categorical \\
 \hline
Thyroid nodule dimensions & Quantitative \\
 \hline
\end{tabular}
\label{tab:DataTypes}
\end{table}

\subsubsection{Demo: predicting using the FPT}
\begin{figure*}[htbp]
    \centering
    \setkeys{Gin}{width=\linewidth} 
\subfloat[Path in PT to classify the synthetic patient. \label{fig:PatientPT}]{\includegraphics{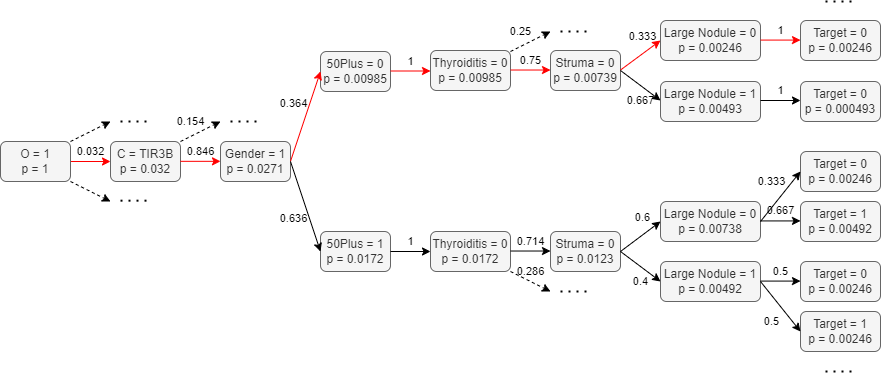} }\hfil
\subfloat[Path in FPT to classify the synthetic patient. \label{fig:PatientPTFIS}]{\includegraphics{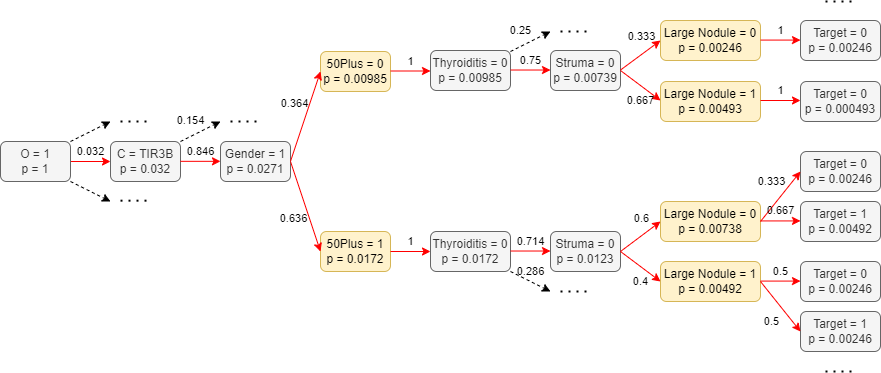} }\hfil

\caption{Results of classifying the same patient using PT vs. FPT. The grey nodes denote crisp variables, where yellow boxes denote fuzzy variables. The path that is taken in the tree to classify the synthetic patient is highlighted by the red arrows. In (b) FPT, all arrows are red as all 'paths' are considered to some degree for the final classification.}
\label{fig:ptvsptfis}
\end{figure*}

A demonstration is presented in order to clarify the decision making process of the proposed FPT method and compare it to the traditional PT. We introduce a synthetic patient with the feature values as presented in Table \ref{tab:syntheticpatient}. The process of classifying the synthetic patient based on PT and FPT is visualised in Figure \ref{fig:PatientPT} and Figure \ref{fig:PatientPTFIS}, respectively. These figures show only the fraction of the tree that is relevant for the classification of this synthetic patient. The red arrows indicate the path taken in the tree to make the classification of the synthetic patient. The probabilities on the arcs represent the transition probabilities to go from one node to a following node. The yellow nodes in the fuzzy tree indicate the nodes that correspond to one of the fuzzy variables. 

\begin{figure*}[htbp]
    \centering 
    \setkeys{Gin}{width=0.35\linewidth} 
\subfloat[We represent the two fuzzy membership sets and membership functions for the linguistic variable 50Plus ('0'=green, '1'=red, yellow area= where the mfs overlap).\label{fig:MFage}]{\includegraphics{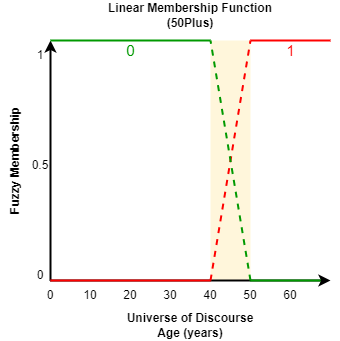} }\hfil
\subfloat[We represent the two fuzzy membership sets and membership functions for the linguistic variable Large Nodule ('0'=green, '1'=red, yellow area= where the mfs overlap). \label{fig:MFsize}]{\includegraphics{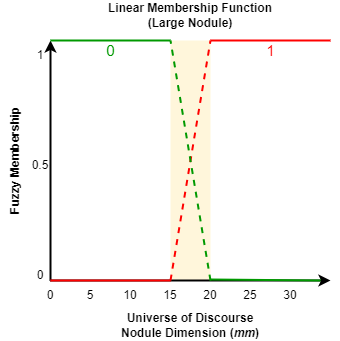} }\hfil
\caption{Linear membership functions for the fuzzy variables (\textit{age} and \textit{nodule dimension}).}
\label{fig:MFs}
\end{figure*}

\begin{table}[htbp]
\centering \small
\caption{Synthetic patient.}
\begin{tabular}{||c|c||}
\hline
\textbf{Feature} & \textbf{Value} \\
\hline
Class & TIR3B\\
\hline
Gender & Female\\
\hline 
Age & 48\\
\hline
Thyroiditis & No \\
\hline
Struma & No \\
\hline
Nodule Dimensions & 18mm \\
\hline
\end{tabular}
\label{tab:syntheticpatient}
\end{table}
We transform two features into linguistic features in order for these features to be modelled as fuzzy variables. The feature for the age of a patient is modelled as the linguistic feature '50Plus' and the feature nodule size is modelled as the linguistic feature 'Large Nodule'. The fuzzy sets are depicted in Figure \ref{fig:MFage} and Figure \ref{fig:MFsize}. In the traditional PT these variables have crisp thresholds. Namely, for the age we make a hard cut-off at the age of 50. A large nodule is defined to be any nodule with dimensions $\geq 20$ millimetre (mm)). 

The design of the tree and the transition probabilities do not change from the PT (Figure \ref{fig:PatientPT}) to the FPT (Figure \ref{fig:PatientPTFIS}). However, what is different is when we make a prediction based on the FPT algorithm. In the case that we aim to predict a data point that has a membership degree in the range $0 < \text{mf} < 1$, we take the weighted average of multiple observations (i.e., we consider multiple paths in the tree). The outcome is a weighted average, where each path is weighed according to its degree of membership in the fuzzy variable. 
The PT (Figure \ref{fig:PatientPT}) classifies the synthesised patient to have a benign nodule ('Target $= 0$') with a probability of $1$. This prediction follows from simply following the path corresponding to the patient's information in the tree, which is denoted by the red arrows. The FPT (Figure \ref{fig:PatientPTFIS}) considers all paths as the patient has membership degrees in the range $0 < \text{mf} < 1$ for both fuzzy variables. Therefore, all arcs leaving the yellow fuzzy nodes are depicted in red. The FPT classifies the synthesised patient with a benign nodule with a probability of $0.427$. \\
This demonstrates the important nuances that incorporating fuzziness in features may bring. Considering a patient aged 48 to belong fully to the group of patients that are 50 years or less, leads to a possible loss of important information. As this patient is very close to the age of 50, the patient should be considered accordingly. Similarly, the FPT considers the feature 'nodule dimensions' to be a fuzzy variable. Whereas in the crisp PT, a large nodule is defined as any nodule with dimensions larger than or equal to 20mm. The FPT implements a linear membership function, herein a nodule with dimensions of 18mm belongs to the class of 'large nodules' to a (membership) degree of $0.80$. The written out calculations, showing the integration of the membership values, to obtain the classification for the synthesised patient as made by the FPT is provided in Equation \ref{fuzzycalc}. 

\begin{equation}\label{fuzzycalc}
\begin{aligned}
    \text{Fuzzy prediction of} & \text{ synthetic patient} = \\
    & 0.2 \cdot \big((0.2 \cdot 1 \cdot 0) + (0.8 \cdot 1 \cdot 0)\big) + \\
    & 0.8 \cdot \big((0.2 \cdot 0.333 \cdot 0) + (0.2 \cdot 0.667 \cdot 1) + \\
    & (0.8 \cdot 0.5 \cdot 0) + (0.8 \cdot 0.5 \cdot 1)\big) = 0.427.
\end{aligned}
\end{equation}


\subsection{A chronic kidney disease case study}
Chronic Kidney Disease (CKD) is a condition in which the kidneys are damaged and cannot filter blood the way that they should. Because of this, excess fluid and waste from blood remain in the body and may cause other health problems, such as heart disease and stroke. Prevalence and incidence of CKD have almost doubled in the past two decades \cite{xie2018analysis}. We aim to predict whether patients will progress to End Stage Renal Disease (ESRD), which is the most advanced and harmful stage of kidney disease. In this stage, patients often need dialysis and/or a kidney transplant. Dialysis is a procedure to remove waste products and excess fluid from the blood when the kidneys stop working properly. Thus, it is a therapy that replaces the normal blood-filtering function performed by the kidneys. Determining the probability of kidney failure may be useful for the communication between the clinicians and patients, furthermore, it will be leading in the triage and management of nephrology referrals and the timing of dialysis and kidney transplants. Reliable and interpretable prediction tools are needed to identify patients with CKD that are at greater risk of developing towards ESRD. The progression is often hard to predict, as the disease does not progress in the same rate for all patients. Moreover, it is said that a substantial portion of patients suffering from CKD do not follow a predictable pattern of disease progression.
\subsubsection{The kidney disease data set}
This study includes a data set that is the result of a multi-centre prospective study pooling data from 46 established hospitals and clinics located in the EU. 
The studies have been conducted for a total of 20 years, starting in 1995 and continuing up to 31 December 2015 \cite{10.1093/ndt/gfy217}. The data set is of tabular form and contains the records of 3.278 patients, each consisting of 131 different features.
Table \ref{tab:DataTypes2} (Appendix \ref{figuresCKD}) presents an overview of the relevant features and their types for this case study. 
From the 3.278 records, a total of 679 have been excluded. 
A number of 236 duplicate patients have been removed, as well as 242 patients that had Glomerular Filtration Rate (GFR) scores that are no indication for CKD. 
Furthermore, 202 patient files have been excluded due to missing information about the levels of potassium (kalium) and/or protein present in the urine. 
This leads to a final data set containing the records of 2.599 patients with CKD. 
For creating, training and testing the model, only complete patient profiles have been used. 
However, the proposed methods can be very useful in making predictions to support the clinical decision processes for patients with missing information. 
The FPT algorithm can be used to create counterfactual statements to test out different scenarios (i.e., different possible values for the specific variable) and evaluate these scenarios for its risks, given the patient information that we know for certain. 
All study participants have signed an informed consent. 

A schematisation of the probability tree is presented in Figure \ref{fig:probabtreeCKD} (Appendix \ref{figuresCKD}), and consists of features that contain general patient information (age, diabetes), features based on blood tests values (serum creatinine, anemia (hemoglobin), GFR), and features based on urine testing values (proteinuria, phosphate). 
On average, the tree will have about five data points in each total realisation. 
However, some patient 'profiles' will be more common than others. 
This number is determined by dividing the total number of patients by the number of possible combinations of feature values: $\frac{2599}{4 \cdot 2^{7}} = 5.08$.

Fuzzy sets are created to represent the variables: serum creatinine, anemia, proteinuria, hyperkalemia, phosphate and the age. 
The aim of creating these fuzzy sets is to create a gradual boundary between the positive and negative class. 
Each variable is represented using two fuzzy sets, all of which are visualised in Figure \ref{fig:fuzzyCKD} (Appendix \ref{figuresCKD}).

\section{Results} \label{results}
In this section we apply our knowledge-based PT and FPT for the analysis of the two clinical data sets. The experimental results that have been achieved by our proposed FPT method are presented, alongside results for two other interpretable benchmark models. Namely, Logistic Regression (LR) and the Decision Tree (DT) model. 

We implement the method of bootstrapping to create multiple data sets based on resampling with replacement. 
On forehand, the data set is split into train and test partitions that are assured to be representative of the data set with the use of stratification. 
The performance metrics in terms of accuracy, specificity, sensitivity, and precision along with corresponding confidence intervals (95\%) are documented. 
The predictions are made using a default threshold, which is set at a probability of 0.50. 
The predicted probability is compared to this threshold, when it is lower than the threshold the predicted classification will be the negative class, when it is higher or equal to the threshold it will be classified as the positive class. 
This threshold can be adjusted according to the situation. For example, it may be increased in times of low capacity when only the most urgent cases can be assessed, i.e., in situations where prioritisation is required. 
\subsection{Results: thyroid nodule case}
The performance of the PT and FPT for the thyroid nodule case are presented in Table \ref{tab:FPTperformance}, similar results are observed for both trees.
This is due to the similarity of the two models, as there are only variables that are represented as fuzzy in the FPT, that are considered using crisp thresholds in the PT. 
Furthermore, the models base their predictions on the same tree, the difference is in the handling of the fuzzy variables (age and nodule dimensions). 
The FPT slightly outperforms the PT across all performance metrics. The biggest difference is in the sensitivity (FPT scores 1 percent point higher), which is evidently the most important performance metric in a medical case study. As this means that the probability that a positive case is in fact positive (i.e., correctly predicting the disease). 
Furthermore, the increase in the lower bound of the CI for the precision of the FPT indicates that it performs more steadily in predicting the disease. 
The increase in performance, although slight, implies that by modelling variables as fuzzy, the FPT is better able to capture the medical scenario. 


\begin{table}[htbp]
\centering \small
\caption[Performance metrics (95\% C.I.) of PT and FPT in Thyroid nodule case study.]{PT and FPT implementation results in Thyroid nodule case study on accuracy, specificity, sensitivity and precision (95\% Confidence Intervals (CIs)). Based on 1000 bootstrapped data sets. Threshold $= 0.50$.}
\begin{tabular}{||c|c|c|c|c||}
\hline
 & \textbf{\begin{tabular}[c]{@{}c@{}}Accuracy\\ (95\% CI)\end{tabular}} & \textbf{\begin{tabular}[c]{@{}c@{}}Specificity\\ (95\% CI)\end{tabular}} & \textbf{\begin{tabular}[c]{@{}c@{}}Sensitivity\\ (95\% CI)\end{tabular}} & \textbf{\begin{tabular}[c]{@{}c@{}}Precision\\ (95\% CI)\end{tabular}} \\ \hline
PT & \begin{tabular}[c]{@{}c@{}}96.8\%\\ {[}93.1 -- 99{]}\end{tabular} & \begin{tabular}[c]{@{}c@{}}98.2\%\\ {[}94.6 -- 100{]}\end{tabular} & \begin{tabular}[c]{@{}c@{}}82.2\%\\ {[}55.6 -- 100{]}\end{tabular} & \begin{tabular}[c]{@{}c@{}}84.1\%\\ {[}58.3 -- 100{]}\end{tabular} \\ \hline
FPT & \begin{tabular}[c]{@{}c@{}}96.9\%\\ {[}94.1 -- 99{]}\end{tabular} & \begin{tabular}[c]{@{}c@{}}98.3\%\\ {[}95.7 -- 100{]}\end{tabular} & \begin{tabular}[c]{@{}c@{}}83.2\%\\ {[}55.6 -- 100{]}\end{tabular} & \begin{tabular}[c]{@{}c@{}}84.2\%\\ {[}63.6 -- 100{]}\end{tabular} \\ \hline
\end{tabular}
\label{tab:FPTperformance}
\end{table}

The performance achieved by the PT and FPT are compared to two benchmark models that are also inherently interpretable, namely LR and DT. 

The performances based on accuracy, specificity, sensitivity, and precision for both the LR and DT model are presented in Table \ref{tab:LRDTperformancesTN}, with corresponding 95\% confidence intervals. Both the LR and the DT are able to handle continuous predictor variables, therefore, the original variables are used instead of the linguistic variables that have been introduced for the age and dimensions ('50Plus' and 'Large Nodule'). The rest of the variables are either categorical or binary, and thus, are not changed. \\
When comparing the performance of the PT and FPT (Table \ref{tab:FPTperformance}) to the performances of the LR and DT presented in Table \ref{tab:LRDTperformancesTN}, the FPT and LR stand out. The LR performs best in terms of accuracy, specificity, and precision. The lower performance in terms of sensitivity indicates that the LR model is more careful in predicting a patient to have the disease, and therefore, being more generous in labelling patients as benign, which is the majority class. This being the reason that the LR outperforms the FPT, we can argue that the FPT is in fact the most useful model. However, the performances of all models are quite comparable and they struggle in correctly classifying the same set of data points. Namely the data points that represent patients that are in the TIR3A or TIR3B class (classes based on the biopsy). As there are very few data points present in the data set that have a class TIR3A (2 data points) or TIR3B (5 data points) and end up having a malignant nodule, it is not surprising that the models generally struggle with classifying these data points. 

\begin{table}[htbp]
\centering \small
\caption[Performances of benchmark models: LR and DT in Thyroid nodule case study.]{LR (no penalty, LBFGS solver) and DT results in Thyroid nodule case study on accuracy, specificity, sensitivity and precision (95\% Confidence Intervals (CIs)). Based on 1000 bootstrapped data sets.}
\begin{tabular}{||c|c|c|c|c||}
\hline
 & \textbf{\begin{tabular}[c]{@{}c@{}}Accuracy\\ (95\% CI)\end{tabular}} & \textbf{\begin{tabular}[c]{@{}c@{}}Specificity\\ (95\% CI)\end{tabular}} & \textbf{\begin{tabular}[c]{@{}c@{}}Sensitivity\\ (95\% CI)\end{tabular}} & \textbf{\begin{tabular}[c]{@{}c@{}}Precision\\ (95\% CI)\end{tabular}} \\ \hline
LR & \begin{tabular}[c]{@{}c@{}}97.1\%\\ {[}94.1 -- 100{]}\end{tabular} & \begin{tabular}[c]{@{}c@{}}98.5\%\\ {[}95.7 -- 100{]}\end{tabular} & \begin{tabular}[c]{@{}c@{}}82.7\%\\ {[}55.6 -- 100{]}\end{tabular} & \begin{tabular}[c]{@{}c@{}}85.8\%\\ {[}66.7 -- 100{]}\end{tabular} \\ \hline
DT & \begin{tabular}[c]{@{}c@{}}96.3\%\\ {[}93.1 -- 99{]}\end{tabular} & \begin{tabular}[c]{@{}c@{}}98\%\\ {[}94.6 -- 100{]}\end{tabular} & \begin{tabular}[c]{@{}c@{}}79.8\%\\ {[}55.6 -- 100{]}\end{tabular} & \begin{tabular}[c]{@{}c@{}}81.5\%\\ {[}57.1 -- 100{]}\end{tabular} \\ \hline
\end{tabular}
\label{tab:LRDTperformancesTN}
\end{table}

\subsection{Results: chronic kidney disease case}
In this section we apply our knowledge-based PT and FPT models to make predictions for the CKD case. The results obtained by the PT and FPT models should differ more than in the previous case study. This is the result of a larger number of fuzzy variables present in the model, making the PT and FPT models less similar to each other. This will help us in understanding more about the differences in performance of the PT/FPT.

The performances for the PT and FPT in their predictions for the progression of CKD is presented in Table \ref{tab:CKDresults}. 
Comparing the performance based on the reported metrics presented shows that the main difference between the two models lies in the sensitivity scores, where the FPT outperforms the PT by 5.5\%. The overall accuracy is slightly improved for the FPT compared to the PT. However, the PT demonstrates slightly better performance for the specificity metric. As regards to the precision, the two models achieve very comparable scores (difference of 0.2 percentage points in favour of the PT). This indicates that the FPTs increase in sensitivity is not achieved simply by the model being more generous with predicting the positive class. Therefore, the results indicate that the FPT is better able to identify the CKD cases that progress to ESRD in the following two year period (the positive class) when compared to the traditional PT. 
However, it must be noted that both models obtain performance that is generally perceived to be low. The sensitivity and precision scores indicate that the models fall short on predicting the CKD cases that progress to ESRD. Nonetheless, the novel FPT method proves to be an improvement over the traditional PT.

The performances are compared to two interpretable benchmark models: LR and DT. The models are given the same combination of features as the (F)PT models. However, LR and DT are both capable of handling continuous predictor variables, therefore, the continuous values of the variables are used. As opposed to the (fuzzy) binary variables that were created for the PT and FPT, that divided the data points into two divisions using splits. 
The performances for the two benchmark models, based on the four performance metrics, are presented in the second half of Table \ref{tab:CKDresults}. Comparing the performance metrics of all models a similar pattern (as in the previous case study) can be observed. The LR is superior in terms of accuracy, specificity, and precision. However, the FPT achieves the highest score in sensitivity. The LR is most accurate in predicting the majority class, however, it is more careful in predicting the positive class (i.e., the minority) and, thus, misses many cases of the disease while scoring higher on the accuracy metric. The FPT is better capable in identifying the positive class, however, its precision in doing so is substantially lower than what is obtained by the LR. Again, it must be noted that all models are limited in their performance, and may not display the performance desired to assist in a clinical environment. 
\begin{table}[H]
\centering \small
\caption[Performance metrics of PT, FPT, LR, and DT in CKD case study.]{PT, FPT, LR and DT results on accuracy, specificity, sensitivity and precision (95\% Confidence Intervals (CIs) in CKD case study). Based on 250 bootstrapped data sets. Thresholds used $= 0.50$.}
\begin{tabular}{||c|c|c|c|c||}
\hline
 & \textbf{\begin{tabular}[c]{@{}c@{}}Accuracy\\ (95\% CI)\end{tabular}} & \textbf{\begin{tabular}[c]{@{}c@{}}Specificity\\ (95\% CI)\end{tabular}} & \textbf{\begin{tabular}[c]{@{}c@{}}Sensitivity\\ (95\% CI)\end{tabular}} & \textbf{\begin{tabular}[c]{@{}c@{}}Precision\\ (95\% CI)\end{tabular}} \\ \hline
PT & \begin{tabular}[c]{@{}c@{}}76.9\%\\ {[}73.7 -- 80{]}\end{tabular} & \begin{tabular}[c]{@{}c@{}}84.1\%\\ {[}78.4 -- 89.2{]}\end{tabular} & \begin{tabular}[c]{@{}c@{}}56.7\%\\ {[}47.6 -- 66.4{]}\end{tabular} & \begin{tabular}[c]{@{}c@{}}56.1\%\\ {[}49.8 -- 63.2{]}\end{tabular} \\ \hline
FPT & \begin{tabular}[c]{@{}c@{}}77.2\%\\ {[}73.8 -- 80.2{]}\end{tabular} & \begin{tabular}[c]{@{}c@{}}82.5\%\\ {[}77.3 -- 87.3{]}\end{tabular} & \begin{tabular}[c]{@{}c@{}}62.2\%\\ {[}52.9 -- 72.3{]}\end{tabular} & \begin{tabular}[c]{@{}c@{}}55.9\%\\ {[}50 -- 62.2{]}\end{tabular} \\ \hline
LR & \begin{tabular}[c]{@{}c@{}}82.7\%\\ {[}80.6 -- 84.8{]}\end{tabular} & \begin{tabular}[c]{@{}c@{}}93.5\%\\ {[}91.5 -- 95.6{]}\end{tabular} & \begin{tabular}[c]{@{}c@{}}52.2\%\\ {[}45.3 -- 59.4{]}\end{tabular} & \begin{tabular}[c]{@{}c@{}}74.1\%\\ {[}67.9 -- 80.8{]}\end{tabular} \\ \hline
DT & \begin{tabular}[c]{@{}c@{}}74.6\%\\ {[}71.7 -- 77.2{]}\end{tabular} & \begin{tabular}[c]{@{}c@{}}82.2\%\\ {[}78.3 -- 85.6{]}\end{tabular} & \begin{tabular}[c]{@{}c@{}}53.1\%\\ {[}44.7 -- 61.1{]}\end{tabular} & \begin{tabular}[c]{@{}c@{}}51.5\%\\ {[}46.2 -- 56.8{]}\end{tabular} \\ \hline
\end{tabular}
\label{tab:CKDresults}
\end{table}


\section{Interpretability analysis} \label{interpretability}
Although the need for interpretability in machine learning models has been established, there is no concrete and agreed upon definition of what constitutes interpretability \cite{linardatos2020explainable, lipton2018mythos}. Therefore, it was proposed in \cite{lipton2018mythos} to define interpretability by identifying two main categories. Namely, an interpretable model should (1) give insights into how the model works in a way that is understandable to a human, and, (2) reveal potential new knowledge. \\
Besides comparing the (F)PT to LR and DT in terms of performance, the methods should also be compared in terms of interpretability. The interpretability of a regular DT and FPT are both based on a rule system that is induced from the tree. Where the edges are connected by 'AND', meaning that we obtain a rule specifying all conditions in a specific branch of the tree. As the depth of a tree increases, the rule system may become more complicated (the length of each rule will be equal to the depth of the tree, i.e., the number of nodes that are in the total realisation). The FPT has an important advantage over DTs, namely that its outputs are presented in the form of probabilities. The prediction is an estimate of the probability of the event occurring, as opposed to a binary classification that is presented by a DT in classification problems. LRs also present their outcomes as a probability of the event occurring. As a LR tries to maximise its log odds (natural logarithm of the odds) function by optimising the parameters. The log odds can be difficult to make sense of, however, the LR presents its estimates as an odds ratio (i.e., probability of the event occurring), which eases the interpretation of the results. Furthermore, it is able to quantify the amount a prediction will change when a certain feature is increased by a value of 1. As a result, FPTs (and DTs) are more inherently comprehensible for humans, as their outputs can be translated to natural language in the form of rules. Whereas, the inner workings of an LR may not be as inherently comprehensible, however, the outputs it presents are interpretable and it is capable of presenting clear feature 'weights', indicating the effect on the target variable after a change in a certain feature value. 

\subsection{Potential of the FPT method (counterfactuals)}
It is important to note that the full potential of the fuzzy probability tree may be yet to be discovered. As probability trees are causal models that use nodes to represent potential states of a process, therefore, the arcs between nodes indicate the probabilistic transition, but preferably also the causal dependency between two states of the process \cite{genewein2020algorithms}. For the two implementations discussed within this research, the trees did not reflect clear processes wherein every node represents an actual state of the process that is a measurable moment in time. Consequently, it is hard to prove any causality between nodes. Possibly, the full potential of (F)PT will be used when there is a clear temporal aspect in the process that the tree aims to model, and the nodes represent actual states of a process. \\
Furthermore, an important power of (F)PTs, lies in its ability to consider counterfactual statements. Counterfactuals allow for the testing of alternate realities, meaning that a doctor may test hypotheses that represent slight alterations to the factual situation. This is particularly importance for comorbidities \cite{capobianco2015comorbiditycorrelations}. An example question that may be answered with the help of counterfactual statements in the CKD case study is: ``\textit{Given a patient that developed kidney failure and did not take any medication. What would be the probability of this patient developing kidney failure if he/she had in fact taken RASI medication?}''. However, careful consultation with clinicians is necessary in creating such counterfactual statements, as RASI medication may induce hyperkalemia, which is again a potentially deadly condition. Nonetheless, this allows clinicians to reason about how the situation may have unfolded had other courses of action been taken. This ability to think in counterfactual statements is natural to humans and it is what sets out human intelligence from other animals. When implementing (F)PTs, this human way of reasoning can be leveraged in AI methods that are also able to consider large amounts of data. The functionality to support the testing of counterfactuals is yet to be developed for the FPT method.  

\section{A tool for clinicians} \label{tool}
Generally, clinicians lack a background in machine learning, probability theory and statistics. In order to create an effective tool, able to assist in the complex decision making processes of clinicians, its information must be well conveyed. Therefore, a tool is created with the aim of allowing clinicians to interactively obtain relevant and naturally understandable statements. These statements are written in the form of probabilities, obtained from the underlying probabilistic decision tree that has been generated based on past patient records. 

The proposed model does not necessarily present definite classifications to the clinicians, as do conventional machine learning algorithms by simply classifying a nodule as being benign or malignant nodules. Rather, the model presents informative probabilities in which it considers the case of a specific patient, based on information given in by the clinician. In this way providing the clinician with a clear indication on the patient's risk of malignancy. Moreover, the clinician can interact with the tool to represent a specific patient in real-time, by changing or filtering the conditions, changes in the probabilities can be observed. This allows the clinician to produce counterfactual statements, and see what the probability for malignancy would be in an alternate reality. It can useful in testing the effects of certain interventions on the target variable. For instance, counterfactual statements could be leveraged to identify the effects of RASI drugs on the progression of CKD. RASI is medication that reduces the future rate of loss of GFR (key indicator of kidney functioning), however, doctors are careful in prescribing it because RASI medication may induce hyperkalemia, which is again a potentially deadly condition. Therefore, testing alternate scenario hypotheses such as ``\textit{Given that this patient has progressed to ESRD, if this patient had taken RASI, what would then be the probability of progressing to ESRD?}'' can be very valuable. Such statements allow for the integration of counterfactual reasoning, human-like reasoning in terms of the fuzziness in variables and large amounts of data. 
\\

The tool is created in the Python 3 programming language and depends on PyQT5 \cite{willman2021overview}, Pandas \cite{mckinney2011pandas}, NumPy \cite{oliphant2007python}, Simpful \cite{spolaor2020simpful}, and the probability tree algorithms as proposed by \cite{genewein2020algorithms}. 
The tool can be launched in Python 3. 
Although the interface is made for the thyroid nodule implementation, the program is created to allow for customisation towards other purposes. The tool is built according to the Model-View-Controller architecture \cite{burbeck1992applications}, which enhances the readability and re-usability of the code, favouring the creation of similar tools for other purposes. 
\\
When launching the user interface all relevant probabilistic trees and fuzzy sets are created based on a csv file from which it reads the data. When new entries are added to the csv file, these will be considered when creating the tree the next time the tool is launched. 

Access to the scripts for the FPT algorithm and tool is available upon request. 





\section{Discussion} \label{discussion}
Clearly, the future of medicine lies in the use and development of human/AI interactive intelligence systems that could leverage on machine learning techniques that are able to analyse large quantities of data to generate a diagnosis for decision support systems while providing an estimate of the degree of uncertainties. It is emerging that medical practice everywhere should be leveraging on sharing information and data in order to improve general healthcare. Implementing such systems will reduce the variability of healthcare among doctors, hospitals and countries. Deep learning has achieved expert-level expertise in many medical areas \cite{duc2022ensemble, han2022automated, mehnatkesh2023intelligent,barbiero2020graph}, nevertheless, it is essential to aim for interpretability in these systems. We believe that interpretability (IAI) is a cautious and wise ground to develop these classes of medical products. As opposed to the less strict concept of explainability (XAI), which we believe is not enough to gain trust and acceptance in the clinical field. Moreover, several regulations imposed by the EU enforce that interpretability is part of the contract. The GDPR states that: ``\textit{... individuals should not be subject to a decision that is based solely on automated processing (such as algorithms) and that is legally binding or which significantly affects them.}'' \cite{EUGDPR}. Furthermore, regulations that have been drafted by the EU in the AIA \cite{EUAIA} aim to specifically regulate the development and use of AI in safety-critical systems (such as the field of healthcare). Any decision support system active in the clinical field will be subject to a strict set of requirements. Conclusively, interpretability must be part of the deal. \\
The results imply that IAI that mimics human reasoning has a fair shot at supporting clinicians in decision making. Allowing the fuzzy, vague and ambiguous medical variables to have fuzzy boundaries will only improve the model and make it even more human-like, while being able to handle large quantities of data. However, more research is needed to identify for which cases the methods work best. As the proposed methods did not achieve consistent results in the two presented case studies. However, the method is successful in it being inherently interpretable, outperforming traditional probability trees, and a closer resemblance to human reasoning which allows it to consider counterfactual statements in combination with fuzzy variables. 
In the context of nephrology, and CKD in particular, the current prediction models are based on Cox proportional hazard models. 
These models often provide estimates on the future risk of future outcomes for each unit increase of a continuous variable or a specific risk category.
The improvement of interpretable FPT may be useful to reach a better prediction of future risk (for ESRD as well as for cardiovascular and mortality events) regardless of a specific cut-off. 
Such a opportunity my overcome the limitation of generate risk categories that can be approximately, and this is particularly true, for instance, for serum bicarbonate and systolic blood pressure or proteinuria \cite{borrelli2018short}.


\section{Conclusion and future work} \label{conclusion}
A novel interpretable decision making method, based on probabilistic trees and fuzzy set theory, was proposed in this paper. The framework, named FPT, exploits the benefits of incorporating fuzzy sets in a probabilistic tree model to support in clinical environments where variables are often vague and ambiguous. Resulting in an inherently interpretable decision making method that is made insightful to its users with the help of a user interface that has been developed specifically for this purpose. Unlike the majority of decision support algorithms existing in literature that are black boxes and/or belong to the class of XAI, FPTs are human-understandable from the beginning. 

We tested and validated the FPT on two distinct clinical case studies. The first representing patients with thyroid nodules that needed to be classified as benign or malignant. The second case included patients suffering from chronic kidney disease, for whom the risk in the following 2-year period to experience kidney failure needed to be assessed. The FPT was compared against several common interpretable benchmark models. Overall, the FPT achieved the best score in detecting the patients that suffer from the disease (i.e., the highest sensitivity) in both case studies. Furthermore, the FPT consistently outperformed the traditional PT in both case studies. This novel interpretable fuzzy probabilistic decision making approach was therefore shown to be a promising addition to the literature on IAI. Incorporating fuzzy variables into the probabilistic model has shown to better reflect the reality of the vagueness that exists in clinical variables. 

We remark that, although the FPT presents an improvement over PTs and to some extent also over a logistic regression, the performance obtained may not yet be sufficient to serve as a reliable decision support tool to assist in daily clinical decision making processes. In future work we aim to test our methods on a larger cohort, hereby demonstrating an increased performance of the FPT and increasing its reliability.
Furthermore, as a future extension of this work, we plan to research the automation of designing the probabilistic trees using heuristics or modelling it as an optimisation problem. In addition, we plan to extend the FPT to support non-binary target variables. 



\section{Acknowledgement}
PL has received funding from the European Union’s Horizon GODS212020. EA thanks Prof Laura Genga for her guidance.

\bibliographystyle{apalike}  
\bibliography{biblio}
\newpage
\begin{appendices}
\section{Chronic kidney disease case study} \label{figuresCKD}
\begin{table}[ht]
\centering\small
\caption[Relevant features Kidney disease data set.]{Explanation of relevant features.}
 \begin{tabular}{||c | c||} 
 \hline
 \textbf{Description} & \textbf{Type} \\ [0.5ex] 
 \hline\hline
Date of birth & Ordinal \\ 
\hline 
Gender & Binary \\
 \hline 
BMI & Quantitative\\
 \hline
Diabetes & Binary \\
\hline
Smoking & Binary\\
 \hline
CVD & Binary \\
 \hline
Kalium (mmol/l) & Binary \\
 \hline
GFR (mg/mmol) & Binary \\
 \hline
Stadium of GFR & Ordinal \\
 \hline
Serum Creatinine (mg/dL) & Quantitative \\
 \hline
Proteinuria (g/24h) & Quantitative \\
 \hline
Hemoglobine (g/dl) & Quantitative \\
 \hline
Phosphate (mg/dl) & Quantitative \\
 \hline
RASI medication & Binary \\
\hline
Date of visits & Ordinal\\
 \hline 
Death pre-dialysis & Binary \\
 \hline
ESRD & Binary \\
 \hline
\end{tabular}
\label{tab:DataTypes2}
\end{table}
\begin{figure}[H]
    \centering
    \includegraphics[width=\textwidth]{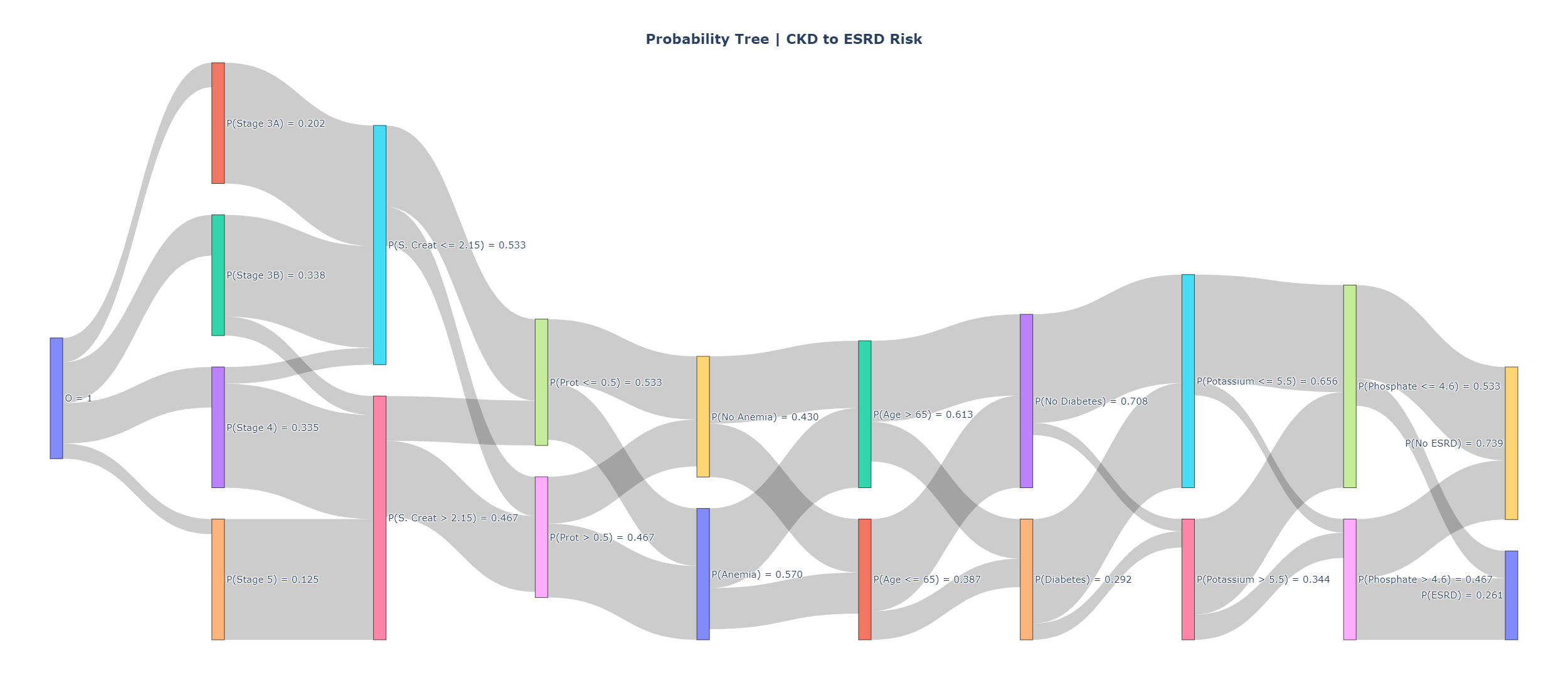}
    \caption[Probability tree developed for CKD data set.]{PT developed for predicting the risk of CKD patients progressing to ESRD. Each arc represents the transition probability from one node (variable) to the next, the size of the arcs are representative of the corresponding transition probabilities.}
    \label{fig:probabtreeCKD}
\end{figure}
\begin{figure}[H]
    \centering
    \setkeys{Gin}{width=0.30\linewidth} 
\subfloat[Fuzzy sets for the variable Anemia (females). \label{fig:anemiaF}]{\includegraphics{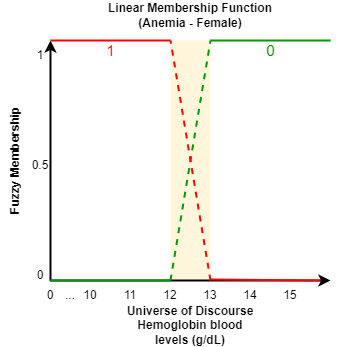} }\hfil
\subfloat[Fuzzy sets for the variable Anemia (males). \label{fig:anemiaM}]{\includegraphics{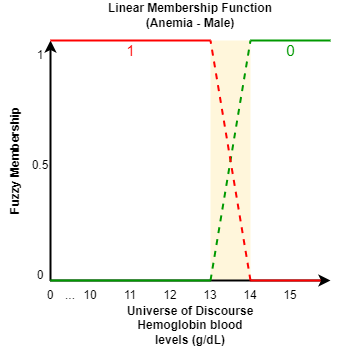} }\hfil
\subfloat[Fuzzy sets for the variable age (65Plus). \label{fig:ageCKD}]{\includegraphics{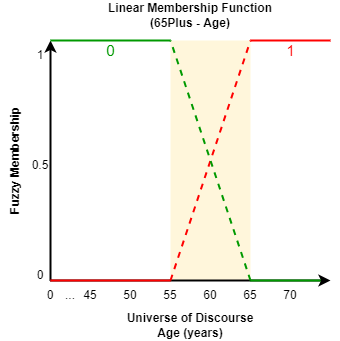} }\hfil
\subfloat[Fuzzy sets for the variable hyperkalemia (serum potassium). \label{fig:hyperk}]{\includegraphics{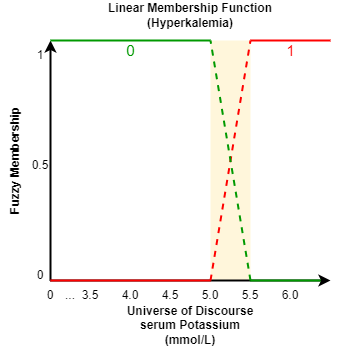} }\hfil
\subfloat[Fuzzy sets for the variable proteinuria. \label{fig:proteinu}]{\includegraphics{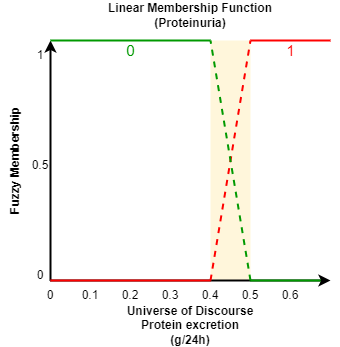} }\hfil
\subfloat[Fuzzy sets for the variable serum creatinine. \label{fig:creat}]{\includegraphics{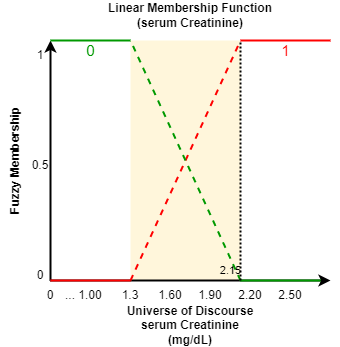} }\hfil
\subfloat[Fuzzy sets for the variable Phosphate. \label{fig:phosp}]{\includegraphics{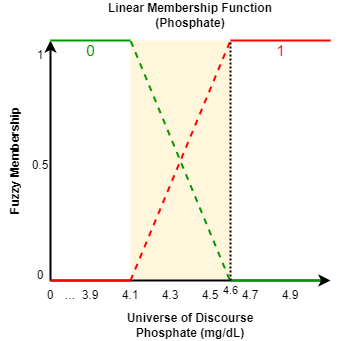} }\hfil
\caption[Fuzzy sets variables CKD case study.]{Fuzzy sets and linear membership functions for the variables: anemia (males and females), age, hyperkalemia, proteinuria, serum creatinine, and phosphate ('0'=green, '1'=red, yellow area= where the sets overlap and thus become fuzzy). \label{fig:fuzzyCKD}}
\end{figure}
\end{appendices}

\end{document}


\maketitle

\section{Membership functions and natural language}

\begin{figure}[!h]
    \centering
    \includegraphics[width=\textwidth]{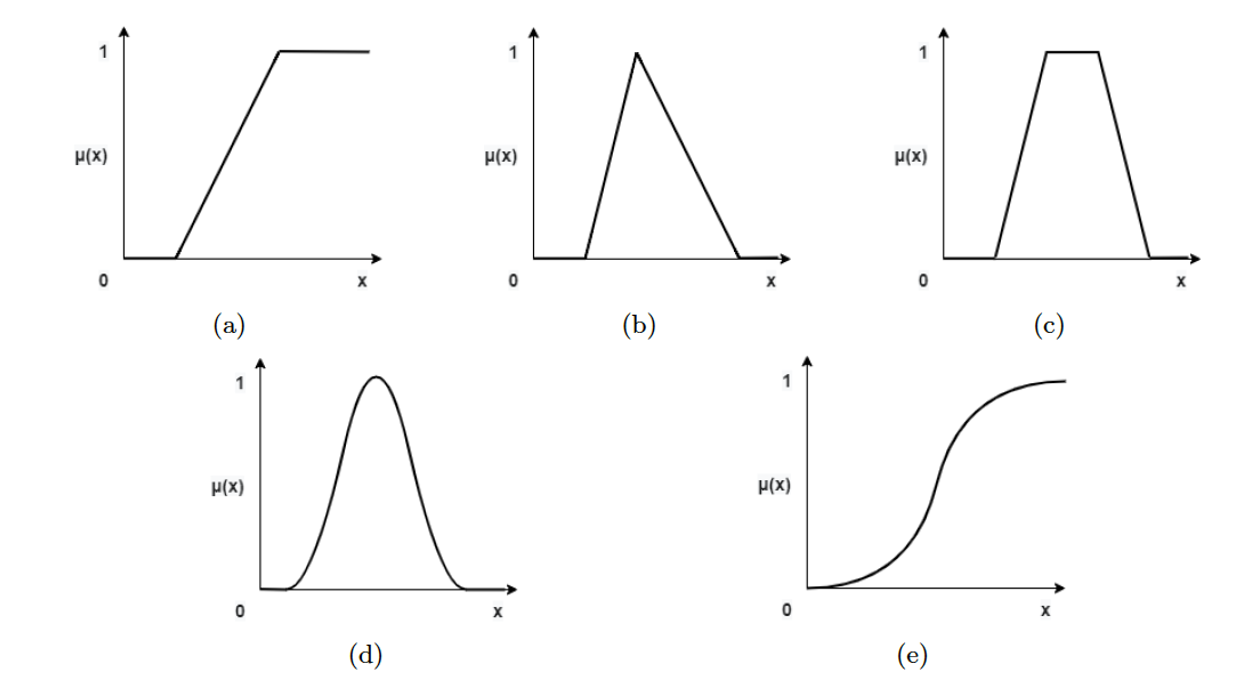}
    \caption{Example of membership functions.}
    \label{fig:fuzzysets}
\end{figure}

Fuzzy sets are mathematical concept that allows for uncertainty and partial membership to a set. 
Differently from traditional sets, where elements can either belong or not belong to a set,  in fuzzy set they can have varying degrees of membership.
Such degree of membership for an element $x \in \Omega$, where $\Omega$ is the universe of discourse, is represented and formalized with a Membership Function (MF) $\mu(x) \in [0,1]$. 
Five examples of membership functions are shown in Figure \ref{fig:fuzzysets}.
The MFs shown in the figure are: (a) rect trapezoid (b) triangular (c) trapezoid (d) Gaussian (e) sigmoid. 

These MFs  can be used to model vague and fuzzy concepts. 
For instance, the body temperature of a patient can be ``high''. 
Although a measured temperature greater than or equal to $38^\circ C$  is usually assumed to be high, also a slightly lower temperature should be considered in a similar way, with a reduced membership degree.
This is the way humans (and medical doctors in particular) generally think.  
That line of reasoning is what a MF like Figure \ref{fig:fuzzysets}(a) represents: the curve is 0 for very low values, slowly increases for medium values, and reaches 1 for high values.
A similar idea, although based on a smoother approach, is represented by the sigmoidal MF in Figure \ref{fig:fuzzysets}(e): the higher the $x$ value, the higher the membership function.

It is possible that the fuzzy concept to be modelled represents a basic condition, e.g., a basal interval of values in non-pathological conditions. 
This idea can be modelled using the MFs in Figures  \ref{fig:fuzzysets}(c) and \ref{fig:fuzzysets}(d).
A MF like \ref{fig:fuzzysets}(b) can be used when the interval to be modelled is not symmetrical.

Linguistic variables can be created using multiple linguistic terms that are typical of medical terminology (e.g., ``low'', ``medium'', ``high'', ``a lot'', ``a few'', ``highly expressed''), where each term is modelled using a dedicated MF, similar to those in Figure \ref{fig:fuzzysets}.